\documentclass{article}

\usepackage{microtype}
\usepackage{graphicx}
\usepackage{subfigure}
\usepackage{booktabs} 

\usepackage{hyperref}

\usepackage[accepted]{icml2018}

\icmltitlerunning{Does Removing Stereotype Priming Remove Bias? }

\begin{document}

\twocolumn[
\icmltitle{Does Removing Stereotype Priming Remove Bias? A Pilot Human-Robot \\Interaction Study}



\icmlsetsymbol{equal}{*}

\begin{icmlauthorlist}
\icmlauthor{Tobi Ogunyale}{goo}
\icmlauthor{De'Aira Bryant}{to}
\icmlauthor{Ayanna Howard}{to}
\end{icmlauthorlist}

\icmlaffiliation{to}{School of Interactive Computing, Georgia Institute of Technology, Atlanta, GA, USA}
\icmlaffiliation{goo}{Department of Computer Science, Georgia State University, Atlanta, GA, USA}

\icmlcorrespondingauthor{De'Aira Bryant}{dbryant@gatech.edu}

\icmlkeywords{Stereotype priming, human-robot interaction, humanoid robot, emotion perception, ICML}

\vskip 0.2in
]



\printAffiliationsAndNotice{} 

\begin{abstract}
Robots capable of participating in complex social interactions have shown great potential in a variety of applications. As these robots grow more popular, it is essential to continuously evaluate the dynamics of the human-robot relationship. One factor shown to have potential impacts on this critical relationship is the human projection of stereotypes onto social robots, a practice that is implicitly known to effect both developers and users of this technology. As such, in this research, we wished to investigate the difference in participants' perceptions of the robot interaction if we removed stereotype priming. This has not yet been a common practice in similar studies.  Given the stereotypes of emotions among ethnic groups, especially in the U.S., this study specifically sought to investigate the impact that robot ``skin color" could potentially have on the human perception of a robot's emotional expressive behavior. A between-subject experiment with 198 individuals was conducted. The results showed no significant differences in the overall emotion classification or intensity ratings for the different robot skin colors. These results lend credence to our hypothesis that when individuals are not primed with information related to human stereotypes, robots are evaluated based on functional attributes versus stereotypical attributes. This provides some confidence that robots, if designed correctly, can potentially be used as a tool to override stereotype-based biases associated with human behavior. 
\end{abstract}

\section{Introduction}

As robotics technology continues to mature, a new generation of robots have emerged with the ability to participate in complex social interactions with humans and other robots [9]. The increasing interest in social robots stems from the increasing potential they have demonstrated to augment and improve services in areas of critical societal importance such as healthcare and education [10, 11]. Yet, for the integration of these robots into society to be successful in the long-term, it is critical that users naturally interact with these social robots without also inheriting the biases prevalent in our own human-human interactions. 

It is well-documented in social psychology that stereotypes play a role in initial human-human interactions [4,5]. One such theme being studied by the human-robot interaction (HRI) community is the effect that stereotypes may have on user perception, trust, and outcomes during human-robot interactions. In particular, researchers have investigated the projections of gender, ethnicity, and age stereotypes on social robots [1,6,12]. Many of these studies provide evidence that stereotype susceptibility impacts human-robot interactions in similar ways to human-human interactions. Psychology researchers have also noted that priming a social category, activating the knowledge of the social stereotype(s) being observed, can automatically elicit stereotype-consistent behaviors. Priming is often interpreted as a demand characteristic to research participants [7]. For example, researchers in [1] asked participants to classify the racial identity of robots in a shooter-bias experiment as either ``White" or ``Black" in addition to administering a set of survey questions regarding attitudes toward and stereotypes about Black and White Americans before initiating their experiment. This represents a classical example of stereotype priming.

In contrast to prior research, this work attempts to assess whether humans perceive robots as different if stereotype priming is not explicitly integrated into the experimental design. As such, we utilize robots of two different skin tones and seek to analyze the effects attributed to not priming human participants with information regarding the stereotypes implicitly applied to the associated ethnic groups. Our hypothesis is that when stereotype priming is withdrawn from human-robot interaction studies, the human interpretations of robot behaviors are based on function rather than existing human stereotypes. By complementing existing work with regards to stereotypes and social robots, we seek to further contribute to understanding the intersection of these two subjects.

\section{Background \& Related Work}

\subsection{Human Stereotypes in Emotion Perception}

Existing stereotypes in emotional perception have been the subject of numerous studies conducted by the sociology and psychology communities. Efforts by [2], [3], [4], and [5] have contributed to the collection of knowledge on the factors that influence and incite the usage of stereotypes in the context of emotional perception.

According to [2], stereotypes are invoked by observers when information used to process emotional intensity is perceived to be ambiguous. For example, if person A were to observe person B yelling expletives, jumping up and down, and clinching their fist, person B would be perceived as angry or mad. Yet, if person A observed person B with just an ambiguous frown on their face, stereotyping person B could influence how person A evaluates their emotion based on their previous encounter. Thus, stereotypes are means to supplement uncertainty in emotion evaluation as a result of ambiguity [2].
Subsequent work has involved investigating the various influences that effect the human evaluation of emotions. Interestingly, studies have collectively observed stereotypes to be the manifestation of a lack of association with people considered different from oneself. Findings show the areas of differences that directly affect emotional evaluation are age [4], gender [3,4], and ethnicity [3,5].

\subsection{Emotional Perception of Robots}

The relationship between robots and humans, with regards to emotion perception, is similar to that of the human-human relationship [6]. In fact, [6] argues that social interactions are universal. Consequently, extensive research has been conducted in the areas of robotic emotional expression and validating the human perception of robotic emotional expression [11,13]. These studies have had very promising results that further support the successful integration of social robots into society. However, the extent to which a projected stereotype onto a robot may affect the human perception of the robot is an associated topic that this work seeks to investigate. 

\subsection{Effects of Stereotype Priming in Humans}

Studies such as [7] and [8] have highlighted a significant correlation between human behaviors and stereotype priming. [7] showed that women who were primed with words relating to the female stereotype or were reminded of their female identity, prior to engaging in the research activity, displayed more stereotype-consistent attitudes toward the fields of arts and math more than women who were primed with male words or reminded of a neutral gender identity. [8] replicated the famed Shih, Pittinsky, and Ambady (1999) priming study with substantive sample sizes and found similar results among participants aware of the targeted stereotypes.  These studies support that participants in such research studies are susceptible to stereotype priming while engaging in a research activity. When not primed, or made of aware of said stereotype, the human behavior was unaffected. This supports the need for critical evaluation of experimental design and methodology in preparing studies within the HRI community. As stereotype priming can affect study results, researchers should work to assure that such biases do not affect research validity. 

\section{Methodology}

\subsection{Robotic Platform}

The ROBOTIS Darwin-Mini Humanoid Robot model was selected for its ability to convey similar human movements in the arms, hips, and lower body. The Darwin-Mini stands 10.6 inches high, weighs 5.0 pounds, and possesses 20 degrees of freedom. Two ROBOTIS Darwin-Mini robots were used for the experiment, a black and white model as seen in Figure 1 and Figure 2, respectively. Similar to the study in [1], we selected the different pigments of the two platforms for their correlation to stereotypes typically derived based on differences in the skin tone of humans. 

\subsection{Survey Design}

A Google survey was designed to conduct the experiment. Participants were asked a set of demographic questions that included gender, ethnicity, education level, and date of birth. After completing the demographic survey, the participants were then asked to identify the different emotions associated with various robot expressive behaviors. 

\begin{figure}[ht]
\begin{center}
\centerline{\includegraphics[width=2.6in,height=1.1in]{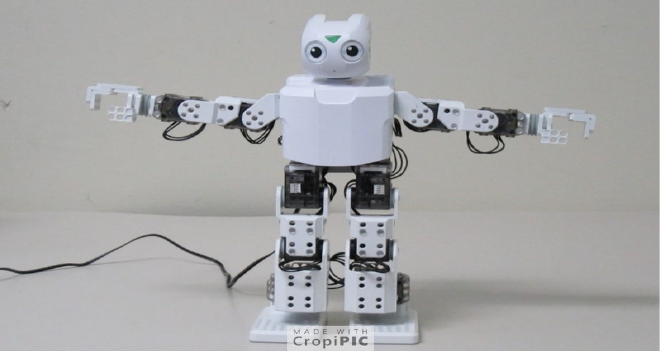}}
\caption{White ROBOTIS Darwin-Mini robot.}
\label{white-robot}
\end{center}
\vskip -0.2in
\end{figure}

\begin{figure}[ht]
\begin{center}
\centerline{\includegraphics[width=2.6in,height=1.1in]{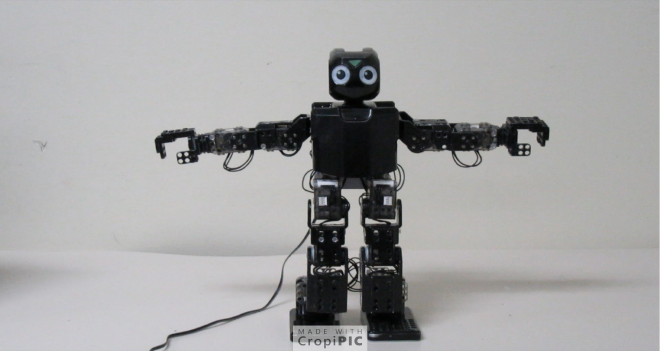}}
\caption{Black ROBOTIS Darwin-Mini robot.}
\label{black-robot}
\end{center}
\vskip -0.2in
\end{figure}

\subsection{Emotion Gesture Set}

To ensure that participants were able to recognize emotions, we utilized a gesture set, as defined in [14], that depicted the positive emotion of happy and the negative emotion of angry. The following variations of these emotions were defined in our final gesture set: happiness: Ha1, Ha2; angry: M1, M2; and the neutral gesture referred to as Non. A description of each gesture in our set can be found in Table 1 along with the gesture's label and the intended emotion.

\begin{table}[t]
\caption{Description of Robot Emotion Gestures.}
\label{gesture-table}
\vskip 0.05in
\begin{center}
\begin{tiny}
\begin{sc}
\begin{tabular}{|p{0.7cm}|p{6.8cm}|} 
\toprule
Gesture & Description \\
\midrule
Ha1      (Happy)    & The Darwin-Mini spreads its arms out to its side with its elbow joints alternating between 25 and 45 degrees. At the same time, the Darwin-Mini's hips are swaying modestly back and forth.
 
\\
Ha2.     (Happy)    & The Darwin-Mini brings its arms out in front and has its elbows joints mimicking a clapping motion. The Darwin-Mini's knees were also alternating, slightly, back and forth.

\\
Non.     (None)    & The Darwin-Mini has it arms out to its sides and does not move from that stance for the duration of the video. 

\\
M1                   

(Angry)    & The Darwin-Mini has its arms out to the side. The right elbow joint is angled at 90 degrees while the left elbow joint is angled at 270 degrees. The arms alternate rapidly up and down. After about 3 seconds, the left arm goes down to the robot's side. Simultaneously, the right arm  waves up and down rapidly.

 \\
M2                   

(Angry)     & The robot initially mimics the motion of pounding a fist into the opposite hand; then, it brings its arms to its sides to mimic the motion of punching its fist into each other. \\
\bottomrule
\end{tabular}
\end{sc}
\end{tiny}
\end{center}
\vskip -0.1in
\end{table}

\section{Experimental Design}

Participants were first given a demographic survey prior to being directed to the research activity. This demographic information was collected to determine if trends would arise between similar individuals when analyzing the robot's emotional state. Upon completing the demographic information, participants were shown a series of videos to evaluate. Each video contained a recording of the Darwin-Mini performing an expressive behavior. After the participant watched a video, they were asked to identify which emotion they perceived the robot to be expressing by selecting a single emotion from the following options: (1) happy, (2) sad, (3) angry, (4) confused, and (5) I don't know. Participants were then asked to rate the intensity of the emotion, on a scale of 1 to 5, with 1 being ``low intensity" and 5 being ``high intensity". 

We conducted a between-subject experiment to test our hypothesis. Participants were recruited via Amazon's Mechanical Turk service and consisted of 200 individuals. The experimental procedures were approved by the Institutional Review Board and all participants acknowledged their consent. Two participants did not complete the study and were therefore dropped from all analysis. The remaining sample consisted of 198 participants (123 males, 73 females, and 2 participants who chose not to specify gender) with a mean age of 35.8 years old and a standard deviation of 13.3. The ethnic breakdown consisted of 68.9\% of the participants self-reporting as Caucasian, 16.1\% as Asian or Pacific Islander, 8.3\% as Hispanic or Latino, and 6.7\% as Black or African-American. A total of 198 surveys were analyzed with 98 responses gathered for the white robot condition and 100 responses gathered for the black robot condition.

\section{Results}

\subsection{Overall Analysis of Emotion Perception}

Figure 3 displays the classification accuracy of which participants were able to recognize the emotions intended by each expressive behavior for both Darwin-Mini models in the two conditions. Accuracy was computed as the total number of correct classifications by participants of the intended robot emotional state. Given that stereotypes are means to supplement uncertainty in emotion evaluation as a result of ambiguity [2], the associated accuracy levels should be consistent across both models if our hypothesis holds true. Figure 4 displays the average intensity rating for each emotion across both conditions. Only intensity ratings associated with the correct emotion classification were factored into the intensity averages.

A confusion matrix, figure 5, was calculated to aid in visualizing the rate at which participants were able to recognize the emotions associated with the expressive behaviors. The overall analysis grouped gestures Ha1 and Ha2 into a category ``Happy" and gestures M1 and M2 into a category ``Angry". Responses associated with the gesture Non were disregarded in the analysis of emotion classification. 

Collectively, participants in both conditions were more easily able to recognize the enactments of angry (White robot, Angry: 58.7\%; Black robot, Angry: 64.2\%), over the enactments of happy (White robot, Happy: 50\%; Black robot, Happy: 50.5\%). A t-test was conducted on the overall average intensity ratings and suggested no statistically significant difference between the perception of either robot model's enactment of happy (White robot: Mean = 3.89, SD = 0.83; Black robot: Mean = 3.90, SD = 0.78), (t = 0.087, P = 0.931, d = 0.115), or angry (White robot: Mean = 4.01, SD = 0.80; Black robot: Mean = 4.17, SD = 0.71), (t = 1.645, P = 0.1013, d = 0.097).

\begin{figure}[ht]
\vskip 0.2in
\begin{center}
\centerline{\includegraphics[width=2.6in,height=1.1in]{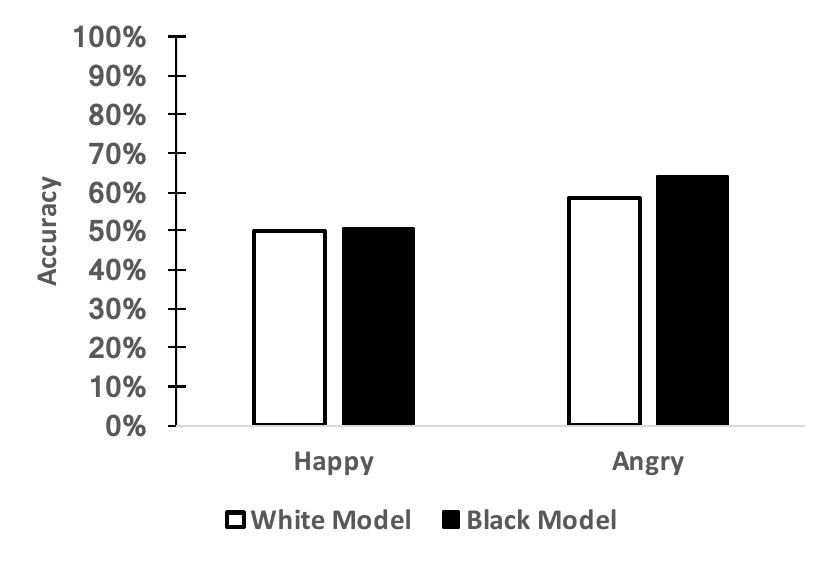}}
\caption{Overall emotion classification accuracy for both conditions.}
\label{overall-accuracy}
\end{center}
\vskip -0.2in
\end{figure}

\begin{figure}[ht]
\vskip 0.2in
\begin{center}
\centerline{\includegraphics[width=2.6in,height=1.1in]{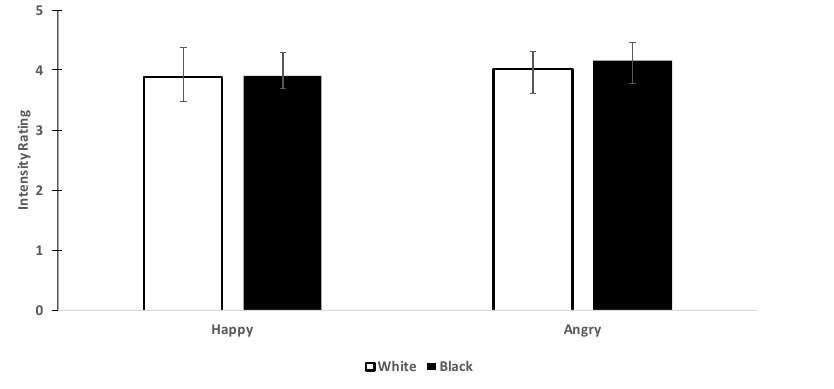}}
\caption{Overall emotion intensity ratings for both conditions.}
\label{overall-intensity}
\end{center}
\vskip -0.2in
\end{figure}

\begin{figure}[ht]
\vskip 0.2in
\begin{center}
\centerline{\includegraphics[width=2.8in,height=1.3in]{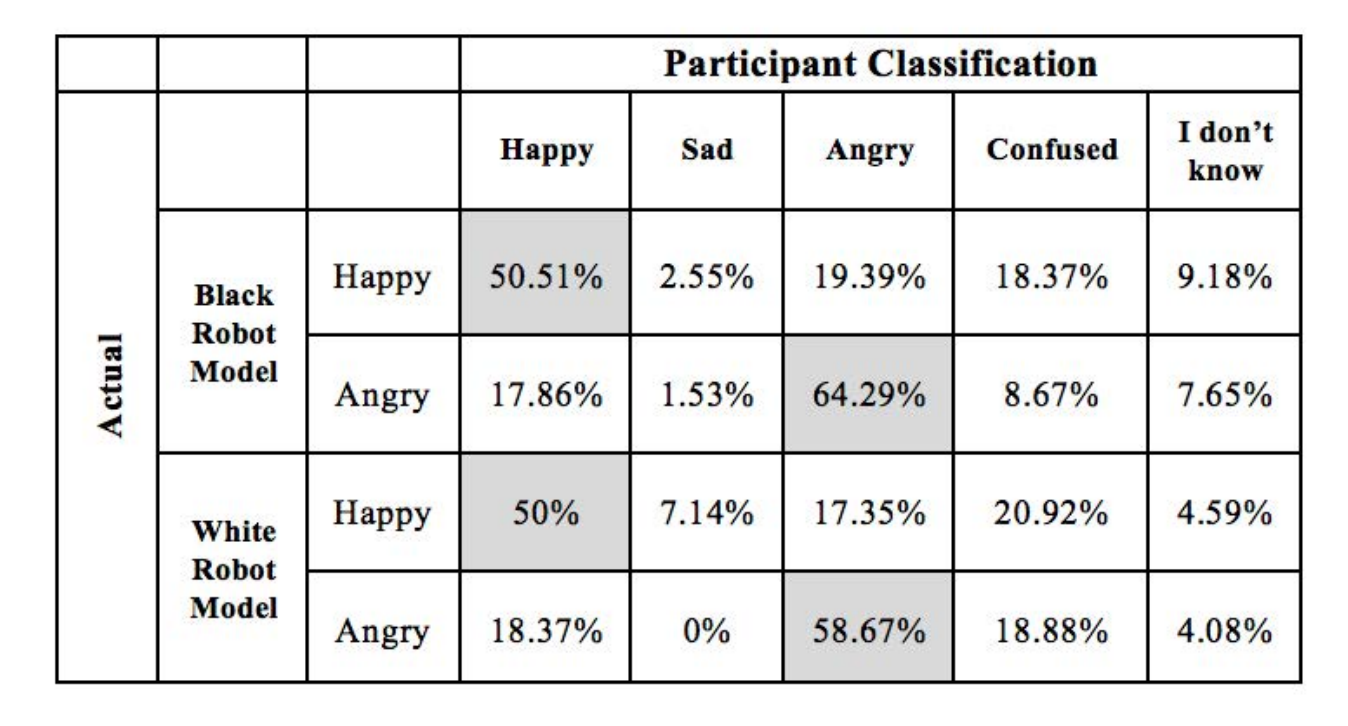}}
\caption{Overall Emotion Classification Confusion Matrix.}
\label{confusion-matrix}
\end{center}
\vskip -0.2in
\end{figure}

\subsection{Gender Effects on Emotion Perception}

Further analysis was conducted to determine if the participant characteristics of gender and ethnicity would affect the emotional perception of the two robot models. Figure 6 shows the classification accuracy rates between male and female participants.

The classification rates for both robot models' enactments of happy were similar for both male and female participants (Male: White robot = 50\%, Black robot = 49.2\%; Female: White robot = 50\%, Black robot = 52.6\%).  The rates for angry were also similar (Male: White robot = 57.9\%, Black robot = 65.8\%; Female: White robot = 60\%, Black robot = 61.8\%).

Figure 7 presents the average intensity ratings with regards to participant genders. Statistical t-tests further revealed that no statistical significance was observed when comparing the emotion intensity ratings of male to female participants or when comparing male to male, or female to female across the two robot models for both the enactments of happy and angry.

\begin{figure}[ht]
\vskip 0.2in
\begin{center}
\centerline{\includegraphics[width=2.6in,height=1.1in]{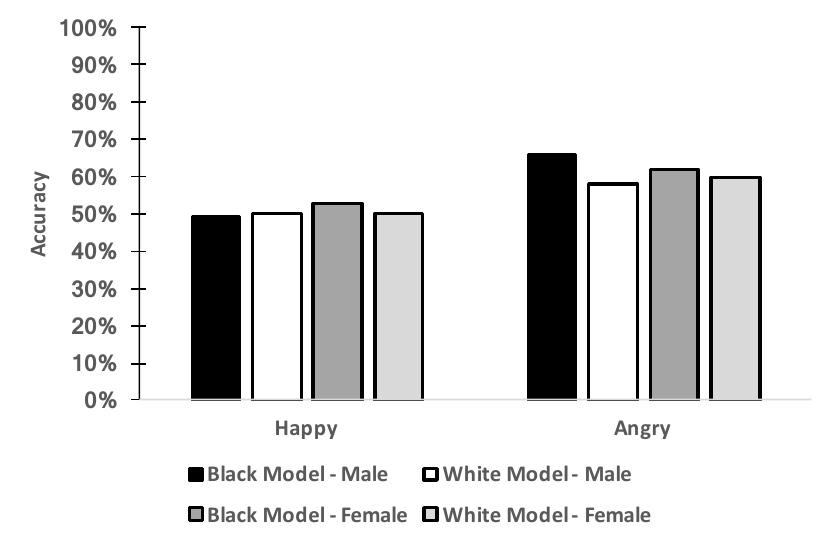}}
\caption{Emotion classification accuracy for both conditions by gender.}
\label{gender-accuracy}
\end{center}
\vskip -0.2in
\end{figure}

\begin{figure}[ht]
\vskip 0.2in
\begin{center}
\centerline{\includegraphics[width=2.6in,height=1.1in]{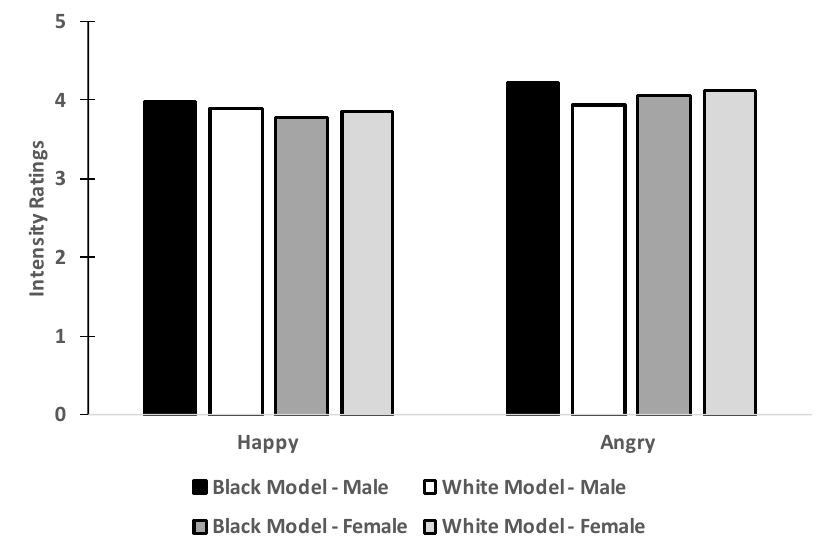}}
\caption{Emotion intensity ratings for both conditions by gender.}
\label{gender-intensity}
\end{center}
\vskip -0.2in
\end{figure}

\subsection{Ethnicity Effects on Emotion Perception}

Figure 8 shows the classification accuracy rates for each of the major ethnic groups while Figure 9 shows the comparison of average intensity ratings between each of the ethnicities. For the black robot model condition, all ethnic groups showed difficulty recognizing the robot's enactments of happy with all recognition rates sitting below 56\%. The enactments of angry were easily recognized by Blacks/African Americans and Caucasians, with ratings of 64.3\% and 61.7\%, respectively. The white robot model condition saw similar results to that of the black robot model condition. However, the enactments of angry saw the ethnic groups of Hispanic/Latino and Caucasian easily able to recognize the expressive behaviors, with ratings of 60\% and 68.1\%, respectively.

The sample size of self-reported Caucasian participants was much larger than any of the other ethnic groups. For this reason, to analyze the average intensity ratings by ethnicity, we compared the data for that of Caucasian participants to the data for that of non-Caucasian participants. A number of t-tests were conducted to make comparisons between the Caucasian participant group and the non-Caucasian group across both robot models and for both the emotions of happy and angry. A series of t-tests were conducted and only 1 test revealed any statistical significance. When considering the intensity ratings of the angry enactments by the black robot, a two-tailed t-test revealed that Hispanic or Latino participants rated the intensity of the expressive behaviors for angry statistically higher than the Caucasian participants. However, due to the extremely small sample size of Hispanic or Latino participants, further research would be needed to draw any solid conclusions. 

\begin{figure}[ht]
\vskip 0.2in
\begin{center}
\centerline{\includegraphics[width=2.6in,height=1.1in]{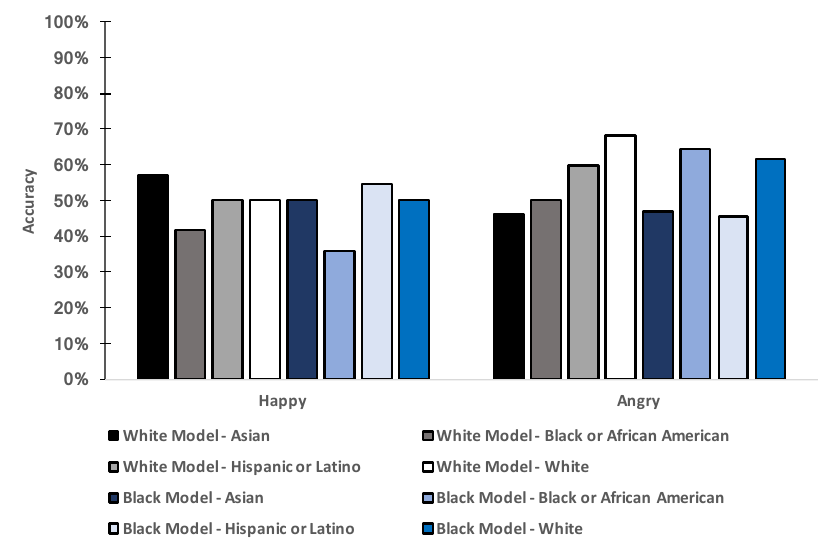}}
\caption{Emotion classification accuracy for both conditions by ethnicity.}
\label{ethnicity-accuracy}
\end{center}
\vskip -0.2in
\end{figure}

\begin{figure}[ht]
\vskip 0.2in
\begin{center}
\centerline{\includegraphics[width=2.6in,height=1.1in]{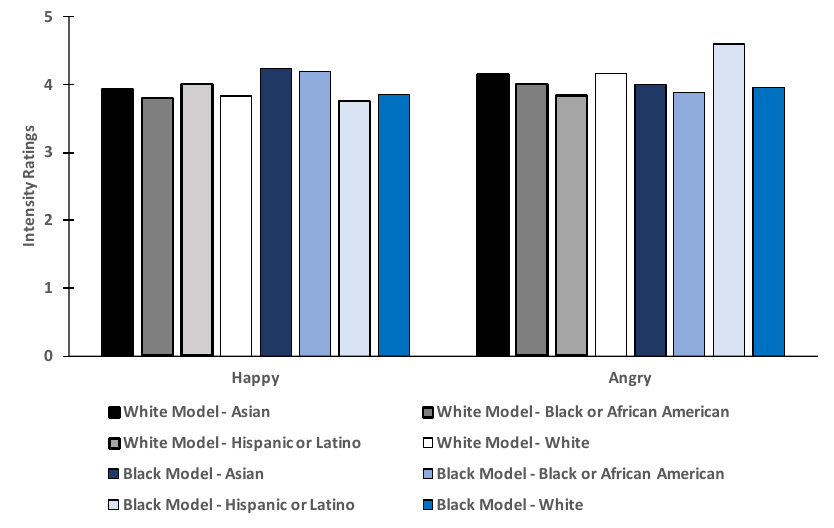}}
\caption{Emotion intensity ratings for both conditions by ethnicity.}
\label{ethnicity-intensity}
\end{center}
\vskip -0.2in
\end{figure}

\section{Discussion}

Our experimental results suggest that participants, when observed as a single collective group, did not evaluate the behaviors of the white and black Darwin-Mini models with any significant difference. When considering gender, the evaluation of robot behaviors between both male and female groups were relatively similar. When analyzing across ethnicity, the same conclusion was reached. These results support the hypothesis that when humans are not susceptible to stereotype priming, their evaluation of robot behaviors is based on functionality rather than the perpetuation of existing human stereotypes.

The experimental design used in [1] incorporates an implicit form of stereotype priming which resulted in the manifestation of an existing human stereotype in the results which contradict the findings we present in this paper. If participants had been influenced by stereotype priming prior to engaging in the research activity, it is likely that existing racial stereotypes would have been evident in our findings. This illustrates that with critical methodological and experimental design, the HRI community can assure that the complex dynamics of the human-robot relationship are evaluated thoroughly without the taint of human biases. 

\section{Limitations \& Future Work}

We must acknowledge the small sample sizes of ethnicities classified as other than Caucasian. This lack of available data can lead to inconclusive comparative results regarding ethnicities when analyzing classification and intensity ratings.

Future work will consist of comparing our current results to a similar experimental design in which participants are primed before being called to evaluate the behaviors and obtaining a larger sample of diverse participants for further analysis. 




\section{References}

[1]	Bartneck, C., Yogeeswaran, K., Ser, Q. M., Woodward, G., Sparrow, R., Wang, S., and Eyssel, F. (2018). Robots And Racism. In Proceedings of the 2018 ACM/IEEE International Conference on Human-Robot Interaction. 196-204.

[2]	Kunda, Z., and Thagard, P. (1996). Forming impressions from stereotypes, traits, and behaviors: A parallel-constraint-satisfaction theory. Psychological review, 103(2), 284.

[3]	Hess, U., Blairy, S., and Kleck, R. E. (2000). The influence of facial emotion displays, gender, and ethnicity on judgments of dominance and affiliation. Journal of Nonverbal behavior, 24(4), 265-283.

[4]	Parmley, M., and Cunningham, J. G. (2014). She looks sad, but he looks mad: The effects of age, gender, and ambiguity on emotion perception. The Journal of social psychology, 154(4), 323-338.

[5]	Hutchings, P. B., and Haddock, G. (2008). Look Black in anger: The role of implicit prejudice in the categorization and perceived emotional intensity of racially ambiguous faces. Journal of Experimental Social Psychology, 44(5), 1418-1420.

[6]	C.D. Martin. The Media Equation: How People Treat Computers, Television and New Media Like Real People and Places [Book Review] (1997). IEEE Spectrum, 34(3), 9-10.

[7]	Jennifer R. Steele and Nalini Ambady. ``Math is Hard!" The effect of gender priming on women's attitudes. Journal of Experimental Social Psychology (2006), 42(4), 428-436.

[8]	Carolyn E. Gibson, Joy Losee, and Christine Vitiello (2014). A replication attempt of stereotype susceptibility (Shih, Pittinsky, \& Ambady, 1999): Identity salience and shifts in quantitative performance. Social Psychology, 45(3), 194-198.

[9]	Anzalone, S. M., Boucenna, S., Ivaldi, S., and Chetouani, M. (2015). Evaluating the engagement with social robots. International Journal of Social Robotics, 7(4), 465-478.

[10]	Liles, K. R., and Beer, J. M. (2015). Rural Minority Students' Perceptions of Ms. An, The Robot Teaching Assistant, as a Social Teaching Tool. In Proceedings of the Human Factors and Ergonomics Society Annual Meeting. 59(1), 372-376. 

[11]	English, B. A., Coates, A., and Howard, A. (2017). Recognition of Gestural Behaviors Expressed by Humanoid Robotic Platforms for Teaching Affect Recognition to Children with Autism-A Healthy Subjects Pilot Study. In International Conference on Social Robotics. 567-576. 

[12]	Chang, R. C. S., Lu, H. P., and Yang, P. (2018). Stereotypes or golden rules? Exploring likable voice traits of social robots as active aging companions for tech-savvy baby boomers in Taiwan. Computers in Human Behavior. 84, 194-210.

[13]	Javed, H., Jeon, M., Howard, A., and Park, C. H. (2018). Robot-Assisted Socio-Emotional Intervention Framework for Children with Autism Spectrum Disorder. In Companion of the 2018 ACM/IEEE International Conference on Human-Robot Interaction. 131-132. 

[14]	Haring, M., Bee, N., and André, E. (2011). Creation and evaluation of emotion expression with body movement, sound and eye color for humanoid robots. In  IEEE Ro-Man. 204-209.

\end{document}